\documentclass[]{fairmeta}

\usepackage{amsmath}
\usepackage{amssymb}

\newcounter{algorithm}
\newcommand{\suppWideTableSetup}{\small\setlength{\tabcolsep}{6pt}\renewcommand{\arraystretch}{1.08}}
\newcommand{\suppTableSetup}{\small\setlength{\tabcolsep}{5pt}\renewcommand{\arraystretch}{1.08}}
\newcommand{\suppDenseTableSetup}{\small\setlength{\tabcolsep}{4pt}\renewcommand{\arraystretch}{1.08}}
\newcommand{\suppBucketTableSetup}{\footnotesize\setlength{\tabcolsep}{3pt}\renewcommand{\arraystretch}{1.04}}

\title{Difficulty-Aware Semantic-ID Optimization for Generative Recommendation}
\renewcommand{\titlelist}{{\fontsize{20}{24}\selectfont\sffamily\bfseries Difficulty-Aware Semantic-ID\\ Optimization for Generative Recommendation}}

\author[1,2]{Xin Yu}
\author[1]{Stephen Li}
\author[1]{Sina Aghaei}
\author[1]{Zifan Zhu}
\author[1]{Jiamu Bai}
\author[1]{Guanjie Huang}
\author[1]{Bo Peng}
\author[1]{Yiyao Liu}
\author[2]{Lingzhou Xue}

\affiliation[1]{Meta}
\affiliation[2]{The Pennsylvania State University}

\date{August 2026}
\correspondence{Lingzhou Xue at \email{lingzhou@psu.edu}}

\abstract{Semantic-ID-based generative recommendation casts retrieval and ranking as autoregressive generation over hierarchical item identifiers. A common recipe is SFT followed by GRPO, yet vanilla GRPO is poorly matched to this tree-structured task. Under the frozen SFT checkpoint, the exact target is absent from the first 16 candidates of the 50-beam constrained ranking for many prompts, and in harder cases none of these candidates enters the target SID branch. This prompt-level diagnostic motivates a training concern: when on-policy GRPO groups are similarly target-missing, item-level rewards may produce weak or degenerate reward variation even if some candidates follow part of the target path. We propose Difficulty-Aware Semantic-ID Optimization (DASO), a tree-aware post-training method that addresses this failure mode as an online rollout-allocation problem. Instead of using fixed difficulty buckets or uniformly injecting ground-truth completions, DASO profiles each current rollout group by prefix-match depth, locates the bottleneck SID levels where candidates leave the target path, and reallocates a bounded portion of the group to prefix-guided completions while retaining raw rollouts for contrast. A SID-prefix reward provides graded credit, while an auxiliary SFT anchor mitigates regression on examples already solved by the SFT checkpoint. On the public benchmarks, DASO improves over MiniOneRec-style GRPO on 11 of 12 metrics and achieves the best result on 9 of 12 metrics; it also improves most level-wise recall metrics on the internal recommendation task. Code: \url{https://github.com/LucasXinYu/DASO}.}

\begin{document}

\maketitle
\noindent{\footnotesize This work was done when Xin Yu was an intern at Meta.}

\section{Introduction}

Generative recommendation offers an end-to-end alternative to conventional recommendation pipelines with separate retrieval and ranking stages. Rather than scoring a fixed candidate set, the model directly generates the target item from user context. Semantic-ID-based generation makes this paradigm practical by representing each item as a short sequence of discrete semantic tokens, turning recommendation into autoregressive item-ID generation \citep{rajput2023recommender}. This idea underlies systems ranging from TIGER-style semantic identifiers to industrial deployments such as OneRec and OneRec-V2, as well as open-source pipelines such as MiniOneRec that combine semantic-ID construction, supervised fine-tuning (SFT), constrained decoding, and recommendation-oriented reinforcement learning \citep{deng2025onerec,zhou2025onerec,zhou2025onerecv2,kong2025minionerec}.

\begin{figure*}[t]
\centering
\includegraphics[width=0.94\textwidth]{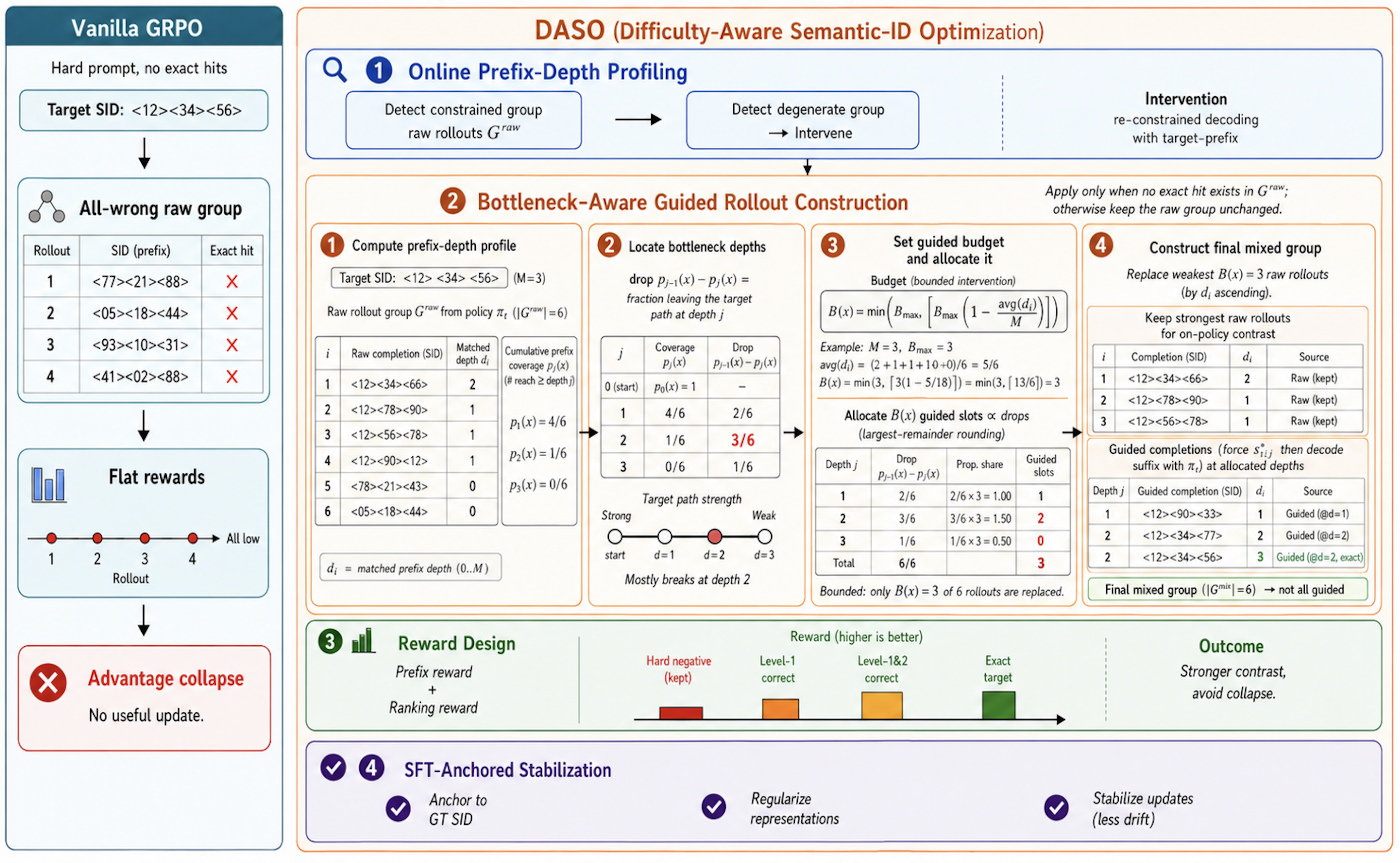}
\caption{Workflow of DASO. For each prompt, DASO samples raw rollouts, builds an $M$-level prefix-depth profile, reallocates a bounded subset to bottleneck-guided completions, and then applies reward normalization and GRPO. The panels are schematic examples of different target-missing regimes: a no-prefix failure and a partial-prefix case handled by the same rule. An auxiliary SFT anchor reduces regression on exact-hit cases.}
\label{fig:workflow}
\end{figure*}

However, this post-training recipe exposes a structural difficulty mismatch. GRPO estimates advantages by comparing multiple rollouts for the same prompt, so it is effective only when the rollout group exhibits meaningful reward variation. In our diagnostic analysis under the frozen SFT checkpoint, the first 16 candidates from the same 50-beam constrained ranking contain no first-token match for 52.5\%--63.9\% of public test prompts. Under the MiniOneRec reward, such groups can collapse to identical zero rewards even when some candidates already match part of the target SID path. In analogous on-policy rollout groups, vanilla GRPO may then fail to distinguish partial-prefix candidates from no-prefix failures, and the group-relative learning signal can become weak or degenerate precisely on the target-missing prompts where post-training should matter most. Figure~\ref{fig:workflow} summarizes the rollout workflow we use to address this issue.

We propose \textbf{Difficulty-Aware Semantic-ID Optimization (DASO)} to address this signal-allocation problem. A useful intervention for semantic-ID GRPO must satisfy three requirements: it should introduce target-path signal on target-missing prompts, preserve enough raw rollouts for group-relative comparison, and avoid hurting examples that the SFT checkpoint already solves. DASO does this through constrained rollout reallocation. At each RL step, it samples a raw group and builds a prefix-depth profile over an $M$-level SID tree. This profile measures how far the current policy follows the target path and identifies the depths where rollouts most often diverge. DASO then reallocates only a bounded subset of the group to prefix-guided completions at those bottleneck depths, while preserving the strongest raw rollouts as on-policy contrastive samples. To limit the distribution shift introduced by guided prefixes, DASO also adds a supervised anchor on the ground-truth SID sequence, improving target-missing cases while reducing regression on exact-hit examples.

Our contributions are summarized as follows:

\begin{itemize}
    \item We diagnose a large target-path coverage gap under the frozen SFT checkpoint: for more than half of the public test prompts, the first 16 candidates from the 50-beam constrained ranking contain no first-token match. This prompt-level gap identifies target-missing regimes where item-level GRPO rewards are prone to weak or degenerate contrast.
    \item We introduce DASO, a tree-aware GRPO post-training method that combines online prefix-depth profiling, bottleneck-aware bounded rollout allocation, SID-prefix credit, and an auxiliary SFT anchor. This design decides both how much guidance to add and which SID depths should receive guidance while retaining raw rollouts for GRPO contrast.
    \item We provide empirical evidence that DASO improves aggregate recommendation quality across two Amazon categories, two Qwen backbones, and an internal four-level SID task. Diagnostic results show the largest bucket-level improvements on initially partial-prefix and no-prefix prompts, and ablations are consistent with the roles of online profiling, SID-prefix credit, guided-budget choice, and SFT anchoring.
\end{itemize}

\section{Related Work}

\subsection{Semantic-ID Based Generative Recommendation}

Recommendation has evolved from collaborative filtering, matrix factorization, and pairwise ranking \citep{koren2009matrix,rendle2009bpr} to sequential models that encode ordered user behavior with recurrent, attentive, convolutional, and self-attentive architectures \citep{hidasi2015session,li2017narm,tang2018personalized,kang2018self,sun2019bert4rec}. More recently, LLM-based recommenders have explored text-to-text prediction, scalable sequence transduction, language-to-item grounding, collaborative semantic indexing, and decoding-time correction or personalization \citep{geng2022recommendation,zhai2024actions,bao2025bistep,zheng2024adapting,bao2024decoding,yu2026exact}. These developments make generation a viable recommendation interface, but many of them still rely on textual grounding, candidate mapping, or decoding-time control rather than directly shaping the semantic-ID rollout distribution.

Semantic-ID-based generative retrieval instead represents each item as a discrete identifier sequence and trains the recommender to generate valid identifiers under constrained decoding, linking recommendation to earlier work on entity and document generation \citep{decao2021autoregressive,tay2022transformer,bevilacqua2022autoregressive}. TIGER shows that hierarchical semantic IDs can support next-item generation and cold-item generalization \citep{rajput2023recommender}; OneRec, its technical report, and OneRec-V2 study end-to-end generative recommendation with preference alignment in industrial settings \citep{deng2025onerec,zhou2025onerec,zhou2025onerecv2}; and MiniOneRec provides an open-source pipeline covering SID construction, SFT, constrained decoding, and GRPO-style training \citep{kong2025minionerec}. OneRec-Think and SIDReasoner further study itemic-token/SID grounding for explicit reasoning \citep{liu2025onerecthink,he2026sidreasoner}. Our work also operates in this semantic-ID setting, but focuses on making the RL stage informative when rollout groups miss the target item or even the correct SID branch.

\subsection{Guided Rollouts and GRPO for Recommendation}

A related direction studies how to make RL effective when successful trajectories are rarely sampled by the current policy. LUFFY mixes on-policy RL with expert-guided traces, BREAD branches rollouts from partial expert anchors, and Prefix-RFT uses demonstration prefixes before autonomous generation \citep{yan2025luffy,zhang2025bread,huang2025prefixrft}. These methods suggest that partial guidance can densify sparse rewards and improve within-group comparisons.

In semantic-ID recommendation, the same exploration issue is shaped by the SID tree. Policy-gradient methods such as PPO and GRPO provide the optimization backbone \citep{schulman2017proximal,shao2024deepseekmath}; Rank-GRPO adapts GRPO to conversational recommendation with rank-level credit \citep{zhu2025rankgrpo}; and V-STAR studies value-guided structured sampling together with Sibling-GRPO for semantic-ID generation \citep{jiang2026vstar}. Relative to this line of work, DASO focuses on a different but related problem: given the current rollout group and the target SID, determine both how much guidance to add and which SID depth to guide, so that GRPO obtains useful reward contrast without replacing policy optimization with full teacher forcing or changing exact-item prediction as the primary objective.

\subsection{Supervised Regularization for Stable Policy Optimization}

RLHF systems commonly initialize from SFT and constrain policy updates with a reference-model KL term to limit drift and reward over-optimization \citep{christiano2017deep,ziegler2019fine}. InstructGPT also shows that supervised or pretraining-style gradients during PPO can reduce regressions on general capabilities \citep{ouyang2022training}. More recent work studies reward-regularized self-distillation beyond pure KL matching and listwise preference optimization in generative models \citep{yu2026pbsd,bai2026listwise}. In recommendation, GenRec similarly combines GRPO with NLL regularization and hybrid rewards to stabilize preference-oriented generation \citep{zou2026genrec}. DASO adopts the same general motivation for a semantic-ID-specific failure mode: prefix-guided rollouts recover signal for groups without exact hits, while supervised anchoring helps reduce regression on examples already solved by the SFT checkpoint.

\section{Preliminaries}
\label{sec:preliminaries}

\subsection{Semantic-ID Tokenization and Generation}

Semantic-ID-based recommendation represents each catalog item $v$ by a short sequence of discrete semantic tokens,
\[
\begin{aligned}
s(v)&=(s_1(v),s_2(v),\ldots,s_M(v)),\\
s_\ell(v)&\in\{1,\ldots,C_\ell\}.
\end{aligned}
\]
Here $M$ is the SID depth and $C_\ell$ is the codebook size at level $\ell$. The SID tokenizer is trained before the recommender is post-trained and then kept fixed. Existing systems construct such identifiers with residual vector quantization or Residual Quantized VAE (RQ-VAE) over item representations \citep{rajput2023recommender,kong2025minionerec}, as well as collaborative- or language-aware vector quantization with balanced non-conflicting assignments \citep{zheng2024adapting}. Ideally, related items share high-level prefixes, so the output space forms a coarse-to-fine semantic tree.

In this tree-structured index, a prefix $s_{1:j}=(s_1,\ldots,s_j)$ corresponds to a node at depth $j$, and $s_{<\ell}$ denotes the tokens before depth $\ell$. During decoding, a prefix trie over valid catalog SIDs constrains the language model to generate only valid item identifiers. Given a user context $x$, we denote the target SID by
\[
s^\star=(s^\star_1,\ldots,s^\star_M).
\]
A generative recommendation policy $\pi_\theta$ predicts a candidate SID autoregressively:
\[
\begin{aligned}
\hat{s}&=(\hat{s}_1,\ldots,\hat{s}_M)
\sim \pi_\theta(\cdot\mid x),\\
\pi_\theta(\hat{s}\mid x)
&=\prod_{\ell=1}^{M}
\pi_\theta(\hat{s}_\ell\mid x,\hat{s}_{<\ell}),
\end{aligned}
\]
where $\hat{s}_{<\ell}$ denotes previously generated SID tokens. The generated SID is then resolved to an item through the fixed SID-to-item index.

Because SIDs are hierarchical, partial correctness is meaningful. We use the prefix-match-depth score $\boldsymbol{m}(\cdot,\cdot)$ to measure the longest matched prefix between a generated SID and the target SID:
\[
\begin{aligned}
\boldsymbol{m}(\hat{s},s^\star)
&=
\max\{j\in\{0,\ldots,M\}:{}\\
&\qquad \hat{s}_{1:j}=s^\star_{1:j}\}.
\end{aligned}
\]
An early mismatch at a high-level SID token routes decoding into an incorrect subtree, making later tokens unlikely to recover the target item. This prefix-level structure is the basis for the DASO prefix-depth profile and SID-prefix reward.

\subsection{Group Relative Policy Optimization}

Group Relative Policy Optimization (GRPO) is a critic-free variant of PPO for language-model post-training \citep{schulman2017proximal,shao2024deepseekmath}. Instead of learning a value function, GRPO estimates advantages by comparing multiple responses sampled for the same prompt. At an RL update step $t$, the rollout policy $\pi_t$ samples a group of $G$ semantic-ID completions:
\[
\begin{aligned}
\mathcal{G}^{(t)}(x)
&=\{\hat{s}^{(1)},\ldots,\hat{s}^{(G)}\},\\
\hat{s}^{(i)}
&\sim \pi_t(\cdot\mid x).
\end{aligned}
\]
Each completion receives a scalar reward $R_i=R(\hat{s}^{(i)},s^\star)$ based on item correctness, generation rank, or SID-prefix progress. GRPO uses the within-group reward statistics as a baseline and computes
\[
A_i
=
\frac{R_i-\operatorname{mean}(\{R_g\}_{g=1}^{G})}
{\operatorname{std}(\{R_g\}_{g=1}^{G})+\delta},
\]
where $\delta$ is a small numerical constant. Thus a completion with reward above the group average receives positive advantage, while a below-average completion receives negative advantage, without requiring a learned critic.

The policy is updated with the standard clipped GRPO/PPO surrogate: token-level likelihood ratios are clipped, and a KL penalty with coefficient $\beta$ keeps the updated policy close to a fixed reference policy $\pi_{\mathrm{ref}}$. We denote the corresponding minimized loss by $\mathcal{L}_{\mathrm{GRPO}}$. DASO keeps this update form and instead changes how the rollout group and rewards are constructed. If all completions receive identical or near-identical rewards, then $A_i\approx 0$ for all $i$, and the policy-gradient signal becomes degenerate. This failure mode motivates the difficulty-aware rollout construction introduced next.

\section{Methodology}
\label{sec:method}

DASO is a target-path-aware rollout reallocation layer for GRPO over semantic-ID trees. Given the current policy's raw rollout group, it decides where a small amount of guidance is most useful. The construction balances three constraints. First, groups without an exact target SID need target-path signal at the right depth: partial-prefix groups need guidance beyond their current matched prefix, while no-prefix groups need help entering the target branch. Second, the final group must retain raw-policy completions so that GRPO still compares candidates produced by the current model \citep{shao2024deepseekmath}. Third, guided prefixes should not amplify regression on examples that the SFT checkpoint already solves \citep{ouyang2022training,zou2026genrec}. DASO uses a deterministic, group-local rule: derive guidance depths from the current prefix-depth profile, use only a bounded intervention budget, keep the GRPO update form unchanged, and add an SFT anchor for stability. The design is related to prior work on partial-guidance rollouts and structured sampling \citep{yan2025luffy,zhang2025bread,huang2025prefixrft,jiang2026vstar}, but here the intervention is chosen from the prompt's current target-path profile rather than from external traces, learned value estimates, or fixed difficulty buckets. Algorithm~\ref{alg:daso} summarizes the procedure.

\begin{table}[tb]
\centering
\refstepcounter{algorithm}\label{alg:daso}
\begin{tabular}{@{}r p{0.86\columnwidth}@{}}
\toprule
\multicolumn{2}{@{}p{0.95\columnwidth}@{}}{\textbf{Algorithm~\thealgorithm: DASO rollout construction and update}}\\
\midrule
\multicolumn{2}{@{}p{0.95\columnwidth}@{}}{\textbf{Input:} prompt $x$, target SID $s^\star$, current policy $\pi_t$, group size $G$, maximum guided budget $B_{\max}$.}\\
\midrule
1 & Sample $G$ valid raw completions from $\pi_t$ under the SID prefix trie.\\
2 & Compute prefix-match depths $d_i$ and the profile $p_j(x)$; the drops $p_{j-1}(x)-p_j(x)$ locate bottleneck depths.\\
3 & If any $d_i=M$, use the raw group unchanged; otherwise set $B(x)$ from the group's average prefix depth.\\
4 & Allocate the $B(x)$ guided slots across depths in proportion to the profile drops.\\
5 & For each selected depth $j$, fix prefix $s^\star_{1:j}$, sample the suffix with $\pi_t$, and replace the weakest raw completions.\\
6 & Compute $R_{\mathrm{DASO}}$, normalize rewards within the final group, and update with $\mathcal{L}_{\mathrm{GRPO}}+\lambda_{\mathrm{sft}}\mathcal{L}_{\mathrm{SFT}}$.\\
\bottomrule
\end{tabular}
\end{table}

\subsection{Online Prefix-Depth Profiling}

At RL update step $t$, all quantities below are computed from a raw group of $G$ valid SIDs sampled by the current policy $\pi_t$ under the prefix trie, following the constrained semantic-ID generation setting used in prior work \citep{rajput2023recommender,kong2025minionerec}. For the $i$-th completion $\hat{s}^{(i)}$, DASO sets $d_i=\boldsymbol{m}(\hat{s}^{(i)},s^\star)$, where $d_i=0$ means that the completion misses the first SID token and $d_i=M$ means that it exactly follows the full target SID. DASO summarizes the group with the cumulative prefix-depth profile
\[
\begin{aligned}
p_j(x)
&=
\frac{1}{G}\sum_{i=1}^{G}
\mathbf{1}[d_i\ge j],
\qquad j=1,\ldots,M.
\end{aligned}
\]
Thus $p_j(x)$ is the fraction of current rollouts that follow the target path to at least depth $j$. A group can miss the exact target while still containing useful partial-prefix evidence: for $M=3$, depths such as $[1,1,0,\ldots]$ indicate a second-level bottleneck, whereas $[0,0,0,\ldots]$ indicates that guidance should enter near the root. The same definition applies to any SID depth: the Amazon experiments use $M=3$, whereas the internal industrial experiment uses $M=4$. Easy, Medium, and Hard buckets are defined only for diagnostic evaluation by applying this profile once under the frozen SFT checkpoint; during training, DASO always uses the current profile from $\pi_t$.

\subsection{Bottleneck-Aware Rollout Allocation}

DASO converts the prefix-depth profile into a final rollout group while satisfying two constraints: the group size remains $G$, and not all raw rollouts are replaced. This keeps the update compatible with GRPO and retains contrastive evidence from the current policy \citep{shao2024deepseekmath}. The construction proceeds in four steps.

\textbf{Step 1: Locate bottleneck depths.} If the raw group already contains the full target SID, DASO leaves the rollout construction unchanged because exact and non-exact candidates already provide group-relative contrast; the prompt still participates in the GRPO and SFT losses. Otherwise, DASO looks for the depths where rollouts first leave the target path. With $p_0(x)=1$, the drop $p_{j-1}(x)-p_j(x)$ is the fraction of raw rollouts that match through depth $j-1$ but fail at depth $j$. Larger drops indicate stronger bottlenecks.

\textbf{Step 2: Decide the total guided budget.} Let $B_{\max}<G$ be the maximum number of raw completions that can be replaced. DASO does not intervene when the raw group already contains the target SID. Otherwise, the budget is larger when the average prefix-match depth is smaller:
\[
\begin{aligned}
B(x)
&=\min\!\Bigg(
B_{\max},\left\lceil B_{\max}
\left(1-\frac{1}{GM}\sum_{i=1}^{G}d_i\right)
\right\rceil
\Bigg).
\end{aligned}
\]
This monotone rule gives more intervention to no-prefix groups and less, but still nonzero, intervention to partial-prefix groups that have not yet produced an exact target SID. We use this simple group-local rule to avoid introducing an additional learned allocator, in contrast to approaches that rely on extra learned value or search modules \citep{jiang2026vstar}. It is deterministic, requires no extra value model, and changes monotonically with target-path progress.

\textbf{Step 3: Allocate the budget across depths.} DASO assigns the guided slots in proportion to the drops $p_{j-1}(x)-p_j(x)$, using largest-remainder rounding to keep exactly $B(x)$ guided completions. This covers both regimes: if all rollouts miss the first token, guidance is concentrated near the root; if several rollouts already match early tokens but none reaches the target item, guidance is concentrated after the matched prefix. Depths that are already frequently matched receive little or no extra guidance, so the method does not treat all target-missing groups as equally hard.

\textbf{Step 4: Construct the final rollout group.} A guided completion assigned to depth $j$ fixes the target prefix $s^\star_{1:j}$ and lets $\pi_t$ decode the remaining suffix. DASO replaces the weakest $B(x)$ raw rollouts, ranked by prefix-match depth, because the strongest raw rollouts already provide the closest on-policy contrast. For guided completions, the forced prefix is treated as conditioning context rather than a sampled action, similar in spirit to prefix-based guidance schemes \citep{huang2025prefixrft,zhang2025bread}; GRPO likelihood-ratio terms are applied only to the policy-generated suffix tokens. The final group therefore combines guided evidence with retained raw completions instead of becoming a purely teacher-forced batch.

\subsection{Difficulty-Aware Reward Design}

After constructing the final group, DASO keeps the MiniOneRec-style recommendation reward \citep{kong2025minionerec} and adds intermediate SID-prefix credit. This follows the general motivation of using richer intermediate preference signals when exact rewards are sparse \citep{zhu2025rankgrpo,jiang2026vstar}. The reward is designed to separate partially correct candidates when exact hits are rare, while keeping exact-item prediction as the primary objective. The binary item reward and prefix-progress reward are
\[
\begin{aligned}
R_{\mathrm{acc}}(\hat{s},s^\star)
&=\mathbf{1}[\hat{s}\text{ resolves to the target}],\\
R_{\mathrm{sid}}(\hat{s},s^\star)
&=\frac{\boldsymbol{m}(\hat{s},s^\star)}{M}.
\end{aligned}
\]
The final reward is
\[
R_{\mathrm{DASO}}(\hat{s},s^\star)=R_{\mathrm{acc}}(\hat{s},s^\star)+\lambda_{\mathrm{rank}}R_{\mathrm{rank}}(\hat{s},s^\star)+R_{\mathrm{sid}}(\hat{s},s^\star).
\]
Here $R_{\mathrm{rank}}$ is the normalized MiniOneRec ranking reward, and its weight $\lambda_{\mathrm{rank}}=0.3$ is inherited unchanged from MiniOneRec. DASO adds the normalized SID-prefix term with unit coefficient in all reported runs. The prefix term supplies graded credit when exact-item rewards are sparse, while the accuracy and ranking terms keep item-level recommendation fidelity central. Since prefix credit is auxiliary and full-item correctness receives separate reward, optimizing a shallow prefix alone cannot maximize $R_{\mathrm{DASO}}$.

\subsection{SFT-Anchored Policy Stabilization}

Prefix-guided completions are designed to recover signal for groups without exact hits, including both partial-prefix and no-prefix groups, but they also shift part of the RL trajectory distribution toward target-prefixed trajectories. This shift is useful for plasticity but risky for stability: the policy may regress on prompts whose target SIDs were already sampled by the SFT checkpoint. Following prior use of supervised regularization to stabilize policy optimization \citep{ouyang2022training,zou2026genrec,yu2026pbsd}, DASO adds a token-level supervised anchor:
\[
\mathcal{L}_{\mathrm{SFT}}
=
-\sum_{\ell=1}^{M}
\log \pi_\theta(
 s_\ell^\star \mid x, s_{<\ell}^\star).
\]
The optimized objective is
\[
\mathcal{L}_{\mathrm{DASO}}
=
\mathcal{L}_{\mathrm{GRPO}}
+\lambda_{\mathrm{sft}}\mathcal{L}_{\mathrm{SFT}}.
\]
Prefix guidance supplies plasticity for groups without exact hits, while the SFT anchor limits drift and mitigates regression on ground-truth SID generation. The anchor is useful because rollout reallocation deliberately changes part of the group distribution toward target-path completions.

\section{Experiments}
\label{sec:experiments}

\subsection{Experimental Setup}

\textbf{Datasets.}
We evaluate DASO on two public Amazon Reviews 2018 categories, \textit{Industrial and Scientific} and \textit{Office Products}, and a proprietary internal user-to-item recommendation dataset. The Amazon experiments follow the common 5-core chronological next-item prediction protocol. The internal experiment uses personalized user contexts to predict target item SIDs; it contains 1.36M main recommendation examples from 319K users, a 1.2M-example SFT pool, 100K GRPO prompts, and about 36K disjoint constrained-generation evaluation prompts.

\begin{table}[tb]
\centering
\setlength{\tabcolsep}{2pt}
\begin{tabular}{@{}p{0.38\columnwidth}p{0.27\columnwidth}p{0.27\columnwidth}@{}}
\toprule
Statistic & Office & Industrial \\
\midrule
\#Items & 3,459 & 3,686 \\
Train / valid / test seq. & 38,924 / 4,866 / 4,866 & 36,259 / 4,532 / 4,533 \\
History mean / median & 3.7 / 3 & 3.8 / 3 \\
Target coverage & 99.4\% & 98.6\% \\
Target Gini / max freq. & 0.528 / 450 & 0.488 / 557 \\
Unique full SIDs / items per SID & 3,444 / 1.004 & 3,670 / 1.004 \\
Used SID codes $L0/L1/L2$ & 88 / 256 / 256 & 48 / 256 / 256 \\
Branch max $a{\to}b$ / $ab{\to}c$ & 66 / 12 & 95 / 47 \\
\bottomrule
\end{tabular}
\caption{Public Amazon dataset and SID statistics. Both datasets use MiniOneRec SIDs with three levels and 256 codes per level.}
\label{tab:dataset_sid_stats}
\end{table}

\textbf{SID construction.}
For Amazon, we use the official MiniOneRec SID assignments: frozen Qwen item-text embeddings are converted into three-level SIDs with 256 codes per level through MiniOneRec's Residual Quantized VAE (RQ-VAE) pipeline, and the mapping is fixed for all methods. For the internal dataset, each item uses a four-level SID with 2048 possible codes per level; the first level covers 1,871 of 2,048 codes and the full target set contains 225,855 unique SIDs. Thus the public setting tests DASO under the released MiniOneRec SID pipeline, while the internal setting tests a deeper and wider industrial SID tree.

\textbf{Training and evaluation.}
We evaluate Qwen2.5-1.5B-Instruct and Qwen2.5-3B-Instruct backbones. Following MiniOneRec, we add SID tokens to the vocabulary, perform full-parameter SFT with cutoff length 512 and batch size 128, and then train GRPO from the shared SFT checkpoint. All methods use constrained decoding with the same valid-SID trie. For GRPO, we use $G=16$, $\beta=10^{-3}$, learning rate $10^{-5}$, and 600 optimizer updates. DASO implements Algorithm~\ref{alg:daso} online: for every prompt-level raw group, it recomputes prefix-match depths, sets $B(x)$ from the group's average prefix depth, allocates guided slots by profile drops, and leaves exact-hit groups unchanged. The MiniOneRec ranking-reward form and $\lambda_{\mathrm{rank}}=0.3$ are inherited unchanged; DASO adds unit-weight SID-prefix credit and uses $B_{\max}=9$ and $\lambda_{\mathrm{sft}}=0.1$. Public evaluation reports HR@5, NDCG@5, and HR@10 with 50-beam constrained decoding. Internal evaluation reports level-wise Top-20 and Top-1 recall. Unless otherwise stated, reported numbers are from single runs. Additional hyperparameters, bucket distributions, and full diagnostic results are provided in the appendix.

\textbf{Baselines.}
We compare with the MiniOneRec SFT checkpoint, MiniOneRec-style GRPO without target-prefix guidance, uniform GT injection, and Sibling-GRPO. MiniOneRec + GT provides a uniform positive-injection comparison without prefix-depth profiling, while Sibling-GRPO provides a tree-structured GRPO comparison. Easy, Medium, and Hard buckets are assigned once using the first 16 candidates from the same 50-beam constrained ranking of the frozen SFT checkpoint. They are used only to partition evaluation examples for diagnostic analysis and are not training-time control variables in DASO.

\subsection{Public Amazon Results}

Table~\ref{tab:main_results} reports the Amazon results. DASO obtains the highest value on 9 of 12 reported metrics and improves over MiniOneRec-style GRPO on 11 of 12 metrics across both datasets and both model sizes. The HR@5 gains are larger on Office Products, from 0.1420 to 0.1639 with the 1.5B backbone and from 0.1313 to 0.1683 with the 3B backbone, than on Industrial and Scientific, where HR@5 improves from 0.1321 to 0.1421 and from 0.1334 to 0.1476. Office also has a larger share of no-prefix Hard prompts, which is consistent with there being more room for recovered training signal in target-missing regimes.

The baselines provide framework-level comparisons. The full DASO configuration obtains higher scores than MiniOneRec + GT on 11 of 12 reported metrics. Because DASO also uses SID-prefix credit, an SFT anchor, bounded guided budgets, and prefix-depth-specific replacement, this comparison should not be read as an isolated test of the allocation rule alone. DASO also obtains higher scores than Sibling-GRPO on 10 of 12 reported metrics; this should likewise be interpreted as a comparison between complete training frameworks rather than a single-factor attribution to profiling. The controlled ablations below provide more direct evidence for individual DASO components. These gains appear in both HR and NDCG, pointing to improved final item ranking rather than only shallow prefix recovery.

\begin{table*}[t]
\centering
\setlength{\tabcolsep}{1.5pt}
\begin{tabular}{@{}llcccccc@{}}
\toprule
\multirow{2}{*}{Backbone} & \multirow{2}{*}{Method} & \multicolumn{3}{c}{Industrial and Scientific} & \multicolumn{3}{c}{Office Products} \\
\cmidrule(lr){3-5}\cmidrule(lr){6-8}
& & HR@5 & NDCG@5 & HR@10 & HR@5 & NDCG@5 & HR@10 \\
\midrule
\multirow{5}{*}{1.5B}
& SFT-only & 0.1035 & 0.0808 & 0.1363 & 0.1231 & 0.0931 & 0.1431 \\
& MiniOneRec & 0.1321 & 0.1084 & 0.1586 & 0.1420 & 0.1172 & 0.1634 \\
& MiniOneRec + GT & 0.1345 & 0.1098 & 0.1601 & 0.1421 & 0.1189 & 0.1683 \\
& Sibling-GRPO & 0.1361 & 0.1128 & \textbf{0.1641} & 0.1500 & 0.1260 & 0.1746 \\
& DASO (Ours) & \textbf{0.1421} & \textbf{0.1183} & 0.1632 & \textbf{0.1639} & \textbf{0.1344} & \textbf{0.1834} \\
\midrule
\multirow{5}{*}{3B}
& SFT-only & 0.1123 & 0.0845 & 0.1332 & 0.1217 & 0.0931 & 0.1471 \\
& MiniOneRec & 0.1334 & 0.1093 & 0.1673 & 0.1313 & 0.1172 & 0.1668 \\
& MiniOneRec + GT & 0.1342 & 0.1102 & \textbf{0.1678} & 0.1456 & 0.1231 & 0.1678 \\
& Sibling-GRPO & 0.1321 & \textbf{0.1119} & 0.1658 & 0.1483 & 0.1267 & 0.1733 \\
& DASO (Ours) & \textbf{0.1476} & 0.1112 & 0.1667 & \textbf{0.1683} & \textbf{0.1331} & \textbf{0.1893} \\
\bottomrule
\end{tabular}
\caption{Amazon Reviews 2018 results with Qwen2.5-1.5B and Qwen2.5-3B backbones.}
\label{tab:main_results}
\end{table*}

\subsection{Internal Industrial Results}

Table~\ref{tab:internal_results} evaluates DASO on the internal four-level SID task, testing whether the prefix-depth logic transfers beyond the three-level Amazon SIDs. The largest change is at the first level: Top-20 lv0 recall increases from 47.21\% to 54.23\%, showing that DASO more often moves generation into the correct coarse branch. This matches the method's emphasis on early bottlenecks. The improvement is not confined to shallow prefixes: final-level lv3 recall also increases from 4.23\% to 4.67\%, and Top-1 recall improves at all four levels. The only exception is Top-20 lv2, where the GRPO baseline is slightly higher; we therefore view the internal result as evidence for stronger branch entry and final SID recovery rather than uniform improvement at every intermediate beam depth.

\begin{table}[!t]
\centering
\begin{tabular}{@{}llcccc@{}}
\toprule
Metric & Method & lv0 & lv1 & lv2 & lv3 \\
\midrule
\multirow{3}{*}{Top-20}
& SFT-only & 46.23 & 11.82 & 4.98 & 3.76 \\
& GRPO baseline & 47.21 & 12.34 & \textbf{5.23} & 4.23 \\
& DASO & \textbf{54.23} & \textbf{12.69} & 5.16 & \textbf{4.67} \\
\midrule
\multirow{3}{*}{Top-1}
& SFT-only & 28.67 & 3.98 & 1.23 & 0.98 \\
& GRPO baseline & 29.12 & 4.03 & 1.23 & 1.02 \\
& DASO & \textbf{29.32} & \textbf{4.21} & \textbf{1.47} & \textbf{1.23} \\
\bottomrule
\end{tabular}
\caption{Level-wise recall on the internal dataset with four-level SIDs. All values are percentages.}
\label{tab:internal_results}
\end{table}

\subsection{Diagnostic Difficulty-Bucket Analysis}

We further analyze results by fixed Easy, Medium, and Hard buckets induced by the frozen SFT checkpoint. These buckets are diagnostic only; DASO training uses the current prefix-depth profile instead of bucket labels. In this diagnostic view, Medium corresponds to partial-prefix prompts and Hard corresponds to no-prefix prompts under the SFT checkpoint. The test sets are dominated by difficult examples: Hard cases account for 63.7\% and 63.9\% of Office Products for the 1.5B and 3B backbones, and 53.3\% and 52.5\% of Industrial and Scientific. This motivates a bucket-level analysis of whether post-training improves target-missing regimes rather than only refining prompts already solved by the SFT checkpoint.

Full bucket-level tables are in the appendix. The largest bucket-level improvements over MiniOneRec are observed on initially Medium and Hard prompts. On Office Products, Hard HR@5 improves from 0.0331 to 0.0624 for 1.5B and from 0.0252 to 0.0521 for 3B over MiniOneRec; Medium prompts also improve substantially. These are the prompt regimes where vanilla item-level GRPO can have weak or degenerate group-relative rewards when analogous target-missing patterns appear in rollouts. Easy prompts, by contrast, are already exact-hit cases under the SFT checkpoint by construction, so SFT-only can remain strong on that subset. Overall, the diagnostic results support the intended role of DASO: recover training signal for partial-prefix and no-prefix prompts while using the SFT anchor to limit regression on prompts that the SFT checkpoint already solves.

\subsection{Ablation Study}

We further isolate four design choices in DASO: online prefix-depth profiling, SID-prefix credit, maximum guided-budget size, and SFT anchoring. Unless otherwise stated, ablations start from the same SFT checkpoint and keep the same GRPO objective, constrained decoding, group size, and training budget.

\textbf{Static difficulty variant.} This variant removes online prefix-depth profiling during RL. Prompts are first assigned to fixed Easy, Medium, and Hard buckets using the frozen SFT checkpoint, and intervention is determined only from the initial bucket rather than from the current rollout group. For the three-level Amazon SIDs, Easy prompts use the raw group unchanged; Medium prompts replace the weakest raw completion with one full-target completion; Hard prompts use a fixed nine-slot recipe, replacing the weakest raw completions with one full-target completion, four completions conditioned on the first two target SID tokens, and four completions conditioned on the first target SID token. In these single-run ablations, Static DASO is below Full DASO on aggregate HR@5/NDCG@5. For 1.5B, it obtains 0.137/0.115 on Industrial and Scientific and 0.154/0.127 on Office Products, compared with 0.1421/0.1183 and 0.1639/0.1344 for Full DASO. For 3B, it obtains 0.138/0.110 and 0.151/0.129, compared with 0.1476/0.1112 and 0.1683/0.1331. The bucket analysis follows the same pattern: Static DASO remains below Full DASO on Medium and Hard prompts, suggesting that the online controller is more effective than this fixed static heuristic.

\textbf{SID-prefix credit and guided budget.} Removing $R_{\mathrm{sid}}$ lowers aggregate Amazon performance in our ablations, and a sensitivity study on Office 1.5B finds that the default $B_{\max}=9$ performs best among $\{5,9,13\}$. This budget study checks sensitivity to the maximum intervention size rather than isolating the full dynamic allocation rule; full tables are in the appendix.

\textbf{DASO without SFT anchoring.} This variant sets $\lambda_{\mathrm{sft}}=0$ while keeping the same prefix-depth allocation and reward. It tests whether guided rollouts alone are sufficient, or whether supervised anchoring is needed to control regression on examples already solved by the SFT checkpoint. Removing the SFT anchor lowers aggregate HR@5 from 0.1421 to 0.139 and from 0.1639 to 0.159 for the 1.5B Industrial and Office settings, and from 0.1476 to 0.143 and from 0.1683 to 0.159 for the 3B settings. After alignment to the 50-beam bucket aggregates, Easy-bucket diagnostics are less uniform, but the aggregate degradation is consistent across all four settings. Prefix-guided rollouts therefore add plasticity, while the SFT anchor helps mitigate overall regression.

\section{Conclusion}

We identify a target-path coverage gap in semantic-ID-based generative recommendation: under the frozen SFT checkpoint, the exact target is absent from the first 16 candidates of the 50-beam constrained ranking for many prompts, and more than half of the public test prompts do not match the first target SID token. Such target-missing regimes can lead to weak or degenerate group-relative rewards during GRPO-style post-training when similar patterns appear in on-policy rollout groups. DASO targets this problem by profiling each current rollout group, allocating prefix-guided completions to bottleneck SID depths, and adding an auxiliary SFT anchor for stability. On the public Amazon benchmarks and an internal industrial dataset, this prefix-depth-aware construction improves aggregate recommendation quality, with the largest bucket-level improvements over MiniOneRec appearing on initially target-missing examples.

\beginappendix

\section{Organization}

This appendix expands the experimental and diagnostic details that cannot fit in the main paper. It is intended to make the evaluation protocol, semantic-ID setting, ablation plan, and bucket-level analysis clear without changing the main method.
Section~\textit{Reproducibility Overview} explains what is fixed across runs and what is included in the public release.
Section~\textit{Dataset and Semantic-ID Details} describes the public and internal recommendation tasks, with emphasis on how the fixed SID trees define the generation space.
Section~\textit{Training and Evaluation Protocol} records the SFT initialization, GRPO-stage hyperparameters, DASO rollout construction, baselines, and metrics.
Section~\textit{Ablation Study Details} reports the controlled ablations.
Section~\textit{Diagnostic Bucket Construction} defines the Easy, Medium, and Hard buckets used only for analysis.
Section~\textit{Full Diagnostic Bucket Results} reports the complete bucket-level Amazon results and explains how they support the main-paper observations.

\section{Reproducibility Overview}

\subsection{Scope of the Appendix}

The main paper focuses on the motivation and aggregate results of DASO. This appendix provides the lower-level experimental information needed to interpret those results. In particular, it records the fixed SID assignments, training stages, decoding constraints, guided-rollout budget, diagnostic bucket construction, controlled ablation variants, and all bucket-level Amazon tables.

\subsection{Run Protocol}

All public experiments use fixed chronological data splits and fixed SID assignments. For each dataset, backbone, and method, we start from the same SFT checkpoint and then run the corresponding GRPO-stage training recipe. Unless otherwise noted, reported numbers are from single runs. This keeps the comparisons focused on the post-training method: differences between MiniOneRec-style GRPO, uniform GT injection, Sibling-GRPO, and DASO are introduced only after the shared SFT initialization.

The implementation fixes random seeds for preprocessing, SFT, rollout sampling, and evaluation scripts. We use the same valid-SID trie for all constrained decoding steps, both during rollout collection and during evaluation. Invalid SIDs are never allowed to be generated, so a method is evaluated on ranking and target selection within the same valid catalog space rather than on syntactic validity. Each reported result is computed from one algorithm run per dataset, backbone, and method setting. Because GRPO rollout collection is computationally expensive, we do not report multi-seed significance tests. Instead, we report the complete protocol and diagnostic breakdowns so that the direction of the gains can be inspected across datasets, model sizes, and initial difficulty levels.

\subsection{Data and Code Availability}

The public experiments use Amazon Reviews 2018 categories with released MiniOneRec-style SID assignments. These public data and SID mappings are sufficient to reproduce the open-source benchmark setting. For the public setting, our code release includes the public-data preprocessing, SID loading, rollout construction, training, and evaluation pipeline where redistribution is permitted.

The internal industrial dataset is proprietary and cannot be redistributed. We therefore report only aggregate task, user, prompt, and SID-space statistics. We omit user-identifying information, raw interaction logs, production catalog metadata, and proprietary data-access code. The reported internal statistics are still useful for interpretation because they describe the depth, width, sparsity, and evaluation scale of the industrial SID-generation task.

\section{Dataset and Semantic-ID Details}

\subsection{Public Amazon Benchmarks}

The public evaluation uses \textit{Office Products} and \textit{Industrial and Scientific}. Both follow a 5-core chronological next-item prediction protocol. For each user, earlier interactions form the history and the next held-out interaction provides the target item. The user history is serialized as the model prompt, and the target item is represented by its fixed semantic ID. Evaluation asks whether the target item appears among the top generated SIDs after constrained decoding.

For these experiments, we use the official MiniOneRec SID assignments. The item textual representations are embedded by a frozen Qwen encoder and then converted into three-level SIDs with 256 possible codes per level through MiniOneRec's Residual Quantized VAE (RQ-VAE) pipeline. We keep the resulting SID-to-item mapping fixed for every method. This is important for fairness: SFT-only, MiniOneRec-style GRPO, uniform GT injection, Sibling-GRPO, and DASO all optimize policies over the same item index and decode with the same trie constraints.

The two Amazon categories are small enough to be reproducible but still expose the difficulty imbalance studied in the main paper. Their target distributions are skewed, and many test prompts are not exact-hit cases under the SFT checkpoint. Therefore, aggregate HR and NDCG results alone do not reveal whether a method improves already-solvable prompts or prompts that initially miss the target path. This motivates the diagnostic bucket analysis reported later in the appendix.

\subsection{Internal Industrial Benchmark}

The internal experiment evaluates whether the same DASO mechanism transfers to a deeper and wider SID tree. Each target item is represented by a four-level SID, with 2048 possible codes at each level. This creates a much larger generation space than the public Amazon setting. The first SID level still has high empirical coverage, while deeper prefixes and full SIDs occupy only a tiny fraction of the theoretical code space. This structure makes early prefix decisions especially important: an error at a high-level token can route the generation into a large incorrect subtree.

The task uses personalized user contexts to generate target item SIDs under constrained decoding. Table~\ref{tab:app_internal_dataset_stats} summarizes the aggregate statistics used for reproducibility and privacy-preserving interpretation. The statistics are reported at the level of examples, users, prompts, and SID coverage rather than raw items or user logs. This allows the internal results to be interpreted as a stress test of DASO on a deeper industrial SID tree while avoiding disclosure of sensitive production information.

\begin{table*}[!tbp]
\centering
\suppWideTableSetup
\begin{tabular}{p{0.54\textwidth}p{0.36\textwidth}}
\toprule
Statistic & Value \\
\midrule
Main recommendation examples & 1,359,570 \\
Users & 319,605 \\
Avg. examples per user & 4.25 \\
Avg. history length & 7.56 \\
SFT train / eval examples & 1,139,832 / 60,168 \\
GRPO train prompts & 100,326 \\
Constrained-generation eval prompts & $\sim$36K \\
Unique target full SIDs & 225,855 \\
SID depth / codebook & $M=4$, $C_\ell=2048$ \\
L1 prefix coverage & 1,871 / 2,048 (91.36\%) \\
L2 prefix coverage & 112,210 / $2{,}048^2$ (2.68\%) \\
L3 prefix coverage & 210,260 / $2{,}048^3$ (0.00245\%) \\
L4 full-SID coverage & 225,855 / $2{,}048^4$ ($\sim 1.3{\times}10^{-6}\%$) \\
\bottomrule
\end{tabular}
\caption{Internal dataset and SID-space statistics. Coverage is computed over the target SID space used by the internal recommendation task.}
\label{tab:app_internal_dataset_stats}
\end{table*}

\FloatBarrier

\subsection{Why SID Structure Matters}

The main methodological assumption behind DASO is not that every dataset has the same SID depth, but that each item identifier forms a path in a fixed semantic tree. A rollout can therefore be evaluated not only by exact item correctness, but also by how far it follows the target path. This property is shared by the three-level public SIDs and the four-level internal SIDs. DASO uses this structure during training through the current prefix-depth profile; the diagnostic bucket analysis uses the same prefix-depth notion only once under the frozen SFT checkpoint.

This distinction matters for interpreting the results. DASO does not need a dataset-specific definition such as ``Easy means first-level match'' or ``Hard means no second-level match''. Instead, it operates on the general depth variable $M$ and can be applied whenever valid item identifiers form a constrained SID tree. The public and internal settings therefore test the same algorithm under different SID depths and branching factors.

\section{Training and Evaluation Protocol}

\subsection{Model Initialization and SFT}

We evaluate Qwen2.5-1.5B-Instruct and Qwen2.5-3B-Instruct backbones. Before SFT, SID tokens are added to the model vocabulary so that the recommender can generate item identifiers directly. We use full-parameter SFT with cutoff length 512, batch size 128, bfloat16 precision, a cosine learning-rate schedule, and early stopping on the validation split. All GRPO-stage methods start from the same SFT checkpoint within each dataset and backbone setting.

The SFT stage provides the initial policy for semantic-ID generation. However, as shown by the diagnostic bucket distributions, SFT does not make every target item appear within the first 16 candidates of the 50-beam constrained ranking. DASO is therefore designed for the post-SFT stage: it keeps the SFT model as the starting point, but changes how rollout groups are constructed and stabilized during GRPO training.

\subsection{GRPO-Stage Training}

All RL-stage methods use constrained decoding with the valid-SID prefix trie. The raw rollout group size is $G=16$. For the standard GRPO components, we use KL coefficient $\beta=10^{-3}$, learning rate $10^{-5}$, and 600 optimizer updates. The KL term uses the SFT checkpoint as the fixed reference policy, following the common post-training setup. Rewards are computed on the final prompt-level group and then normalized within that group before computing group-relative advantages.

\paragraph{Reward computation.}
For a prompt-level group, let $\rho_i\in\{1,\ldots,G\}$ denote the rank position of completion $\hat{s}^{(i)}$ in the constrained generation group, and let
\[
D_G=\sum_{r=1}^{G}\frac{1}{\log_2(r+1)}.
\]
The MiniOneRec rule reward is the exact-item reward $R_{\mathrm{acc}}(\hat{s}^{(i)},s^\star)=\mathbf{1}[\hat{s}^{(i)}\text{ resolves to target}]$. Its rank-aware component is active only when the group contains at least one exact target completion:
\[
R_{\mathrm{rank}}^{(i)}=
\begin{cases}
0, & \text{if no exact target appears in the group},\\
0, & \text{if } \hat{s}^{(i)} \text{ resolves to target},\\
-\dfrac{1}{D_G\log_2(\rho_i+1)}, & \text{otherwise}.
\end{cases}
\]
Thus high-ranked negatives receive larger penalties, while target-missing groups receive no MiniOneRec rank contrast. DASO uses this inherited rank reward with $\lambda_{\mathrm{rank}}=0.3$ and adds $R_{\mathrm{sid}}(\hat{s}^{(i)},s^\star)=\boldsymbol{m}(\hat{s}^{(i)},s^\star)/M$ with unit weight before group-wise reward normalization.

DASO uses a maximum guided replacement budget $B_{\max}=9$ and SFT anchor weight $\lambda_{\mathrm{sft}}=0.1$. For each prompt, DASO first samples the raw group from the current rollout policy. If the raw group already contains the full target SID, DASO does not intervene in rollout construction, but the prompt still contributes to the GRPO and SFT losses. Otherwise, it computes the prefix-depth profile, sets the number of guided replacements from the group's average prefix depth, assigns guided completions to depths where the raw group most often leaves the target path, and replaces the weakest raw completions while keeping the final group size equal to $G$. This keeps the training interface identical to GRPO while changing the information content of the group.

The guided completions are not used as a separate supervised batch. Each guided completion fixes a target prefix and lets the current policy generate the remaining suffix under the same valid-SID constraints. Forced-prefix tokens are treated as conditioning context rather than sampled actions: they are excluded from the GRPO likelihood-ratio and KL terms, and policy-gradient terms are applied only to suffix tokens generated by the current policy. After raw and guided candidates are merged, rewards and group-relative advantages are computed on the final group. Invalid SIDs are ruled out by the prefix trie. Duplicate SIDs within the same prompt-level group are suppressed by the constrained decoding and final-group construction routine, so a duplicated item is not counted as an additional rollout. The purpose is to restore reward contrast inside groups without exact hits, not to replace policy optimization with full ground-truth imitation. The auxiliary SFT loss is optimized in parallel to reduce regression on examples that were already solved after SFT.

DASO is optimized with GRPO rather than Direct Preference Optimization (DPO), but it follows a related post-training principle: preference or reward signals are most effective when paired with constraints that control policy drift. DPO replaces explicit reward-model RL with a stable preference-classification objective \citep{rafailov2023direct}. Recent extensions further show that preference objectives can be adapted to listwise generative feedback \citep{bai2026listwise} and to reward-regularized self-distillation beyond pure KL matching \citep{yu2026pbsd}. These works motivate our stabilization choice at a high level, while DASO addresses a different failure mode: sparse group-relative advantages over a constrained semantic-ID tree.

\begin{table}[!htbp]
\centering
\suppDenseTableSetup
\begin{tabular}{p{0.50\columnwidth}p{0.39\columnwidth}}
\toprule
Training setting & Value \\
\midrule
Backbones & Qwen2.5-1.5B/3B-Instruct \\
Public SID depth / codebook & $M=3$, $C_\ell=256$ \\
Internal SID depth / codebook & $M=4$, $C_\ell=2048$ \\
SFT cutoff length & 512 \\
SFT batch size & 128 \\
SFT optimization & Full-parameter, cosine schedule, early stopping \\
Precision & bfloat16 \\
Raw group size $G$ & 16 \\
Maximum guided replacements $B_{\max}$ & 9 \\
KL coefficient $\beta$ & $10^{-3}$ \\
GRPO learning rate & $10^{-5}$ \\
Ranking reward weight $\lambda_{\mathrm{rank}}$ & 0.3 \\
SID-prefix reward weight & 1.0 \\
SFT anchor weight $\lambda_{\mathrm{sft}}$ & 0.1 \\
GRPO updates & 600 \\
Public evaluation decoding & 50-beam constrained decoding \\
Internal evaluation metrics & Level-wise Top-20 and Top-1 recall \\
\bottomrule
\end{tabular}
\caption{Training and evaluation hyperparameters used in the main experiments.}
\label{tab:app_hyperparams}
\end{table}

\FloatBarrier

\subsection{Baselines}

The SFT-only baseline evaluates the shared supervised checkpoint without any RL post-training. MiniOneRec denotes the MiniOneRec-style GRPO recipe without target-prefix guidance. MiniOneRec + GT injects ground-truth completions uniformly, providing a comparison to positive injection without profiling where the current rollouts leave the target path. Sibling-GRPO is a structured GRPO baseline that uses tree relations among sibling candidates. DASO differs from these baselines by allocating both the number and depth of guided completions according to the current prefix-depth profile of each prompt-level rollout group.

This baseline set provides three complementary framework-level comparisons. Comparing DASO with MiniOneRec shows the effect of adding target-path guidance, SID-prefix credit, and SFT anchoring to the MiniOneRec-style GRPO recipe. Comparing DASO with MiniOneRec + GT contrasts the full DASO configuration with uniform ground-truth injection. Comparing DASO with Sibling-GRPO compares the full DASO framework with a sibling-based tree-structured GRPO variant; because DASO also changes reward design and stabilization, this comparison should not be interpreted as an isolated test of prompt-specific profiling. Component-level ablations in the following sections probe individual design choices more directly.

\subsection{Evaluation Metrics}

For the public Amazon benchmarks, we report HR@5, NDCG@5, and HR@10 under constrained beam search with 50 beams. A generated SID is counted as a hit when it resolves to the target item. HR measures whether the target appears in the top-$K$ candidate set, while NDCG also rewards higher rank positions. These metrics match the standard next-item recommendation protocol used by the compared generative recommendation baselines.

For the internal experiment, we report level-wise Top-20 and Top-1 recall. Level-wise recall measures whether the generated candidate set matches the target SID prefix up to a given depth. For example, lv0 recall evaluates whether the first SID token matches the target branch, while the final level measures full-depth SID recovery. This metric is useful for the internal four-level tree because it separates coarse semantic-branch selection from final item-level identification, matching the prefix-depth view used by DASO during training.

\section{Ablation Study Details}

This section records controlled ablations that probe the roles of online prefix-depth profiling, SID-prefix credit, maximum guided-budget size, and supervised anchoring. Unless explicitly changed, all ablations use the same datasets, SFT checkpoint, SID trie, rollout group size, GRPO hyperparameters, reward definition, and evaluation metrics as the main experiments.

\subsection{Static Difficulty Variant}

The static variant is an ablation baseline rather than the DASO training controller. It removes online prefix-depth profiling from DASO. Before RL training, prompts are assigned to Easy, Medium, and Hard buckets using the first 16 candidates of the frozen SFT checkpoint's 50-beam constrained ranking. During RL, the intervention rule depends only on this initial bucket assignment and is not recomputed from the current rollout group. For the three-level Amazon SIDs, Easy prompts use the raw group unchanged. Medium prompts replace the weakest raw completion with one full-target completion, keeping the other $G-1$ raw completions. Hard prompts use a fixed nine-slot recipe: one full-target completion, four completions conditioned on the target prefix $s^\star_{1:2}$, and four completions conditioned on the target prefix $s^\star_{1:1}$; the remaining suffix tokens are sampled from the current policy under the same SID trie. These nine completions replace the weakest raw rollouts ranked by prefix-match depth, while the seven strongest raw rollouts are retained. This ablation tests whether current-group profiling provides advantages over this one-time difficulty label and fixed intervention recipe.

Table~\ref{tab:app_ablation_aggregate} reports that Static DASO is below Full DASO on all reported aggregate HR@5 and NDCG@5 metrics. Table~\ref{tab:app_ablation_static_bucket} further indicates that the gap is visible on both Medium and Hard prompts, especially for Office Products. These results suggest that, under this tested static heuristic, the current prefix-depth profile provides more useful guidance than a frozen pre-training bucket label as RL changes the policy.

\begin{table*}[!tbp]
\centering
\suppWideTableSetup
\begin{tabular}{llccccc}
\toprule
Backbone & Dataset & Full HR@5 & Static HR@5 & w/o SFT HR@5 & Full NDCG@5 & Static NDCG@5 \\
\midrule
1.5B & Industrial & 0.1421 & 0.137 & 0.139 & 0.1183 & 0.115 \\
1.5B & Office & 0.1639 & 0.154 & 0.159 & 0.1344 & 0.127 \\
3B & Industrial & 0.1476 & 0.138 & 0.143 & 0.1112 & 0.110 \\
3B & Office & 0.1683 & 0.151 & 0.159 & 0.1331 & 0.129 \\
\bottomrule
\end{tabular}
\caption{Aggregate ablation results on Amazon. Static removes online prefix-depth profiling; w/o SFT sets $\lambda_{\mathrm{sft}}=0$.}
\label{tab:app_ablation_aggregate}
\end{table*}

\begin{table}[!htbp]
\centering
\suppDenseTableSetup
\begin{tabular}{llccc}
\toprule
Backbone & Dataset / Bucket & Full & Static & Gap \\
\midrule
1.5B & Office / Medium & 0.1693 & 0.147 & 0.0223 \\
1.5B & Office / Hard & 0.0624 & 0.049 & 0.0134 \\
1.5B & Industrial / Medium & 0.1320 & 0.129 & 0.0030 \\
1.5B & Industrial / Hard & 0.0490 & 0.043 & 0.0060 \\
3B & Office / Medium & 0.2102 & 0.180 & 0.0302 \\
3B & Office / Hard & 0.0521 & 0.039 & 0.0131 \\
3B & Industrial / Medium & 0.1415 & 0.129 & 0.0125 \\
3B & Industrial / Hard & 0.0577 & 0.047 & 0.0107 \\
\bottomrule
\end{tabular}
\caption{Static-difficulty ablation on diagnostic Medium and Hard buckets. Values are HR@5; Gap is Full DASO minus Static.}
\label{tab:app_ablation_static_bucket}
\end{table}

\FloatBarrier

\subsection{Removing the SID-Prefix Reward}

This variant removes $R_{\mathrm{sid}}$ from the DASO reward while keeping online prefix-depth allocation, $B_{\max}=9$, $\lambda_{\mathrm{rank}}=0.3$, and $\lambda_{\mathrm{sft}}=0.1$ unchanged. Table~\ref{tab:app_ablation_nosid} indicates that the full reward is better on all aggregate Amazon metrics. The differences are modest but consistently favor the full reward in these runs, supporting the role of SID-prefix credit as an intermediate signal for target-missing rollout groups rather than as the sole source of improvement.

\begin{table}[!htbp]
\centering
\suppDenseTableSetup
\begin{tabular}{llccc}
\toprule
Backbone & Dataset & HR@5 & NDCG@5 & HR@10 \\
\midrule
\multicolumn{5}{l}{Full DASO} \\
1.5B & Industrial & 0.1421 & 0.1183 & 0.1632 \\
1.5B & Office & 0.1639 & 0.1344 & 0.1834 \\
3B & Industrial & 0.1476 & 0.1112 & 0.1667 \\
3B & Office & 0.1683 & 0.1331 & 0.1893 \\
\midrule
\multicolumn{5}{l}{w/o SID Reward} \\
1.5B & Industrial & 0.138 & 0.113 & 0.161 \\
1.5B & Office & 0.155 & 0.129 & 0.179 \\
3B & Industrial & 0.143 & 0.111 & 0.162 \\
3B & Office & 0.158 & 0.131 & 0.187 \\
\bottomrule
\end{tabular}
\caption{Ablation of the SID-prefix reward on Amazon. The w/o SID Reward variant removes $R_{\mathrm{sid}}$ and keeps the same online rollout allocation, SFT anchor, and remaining reward terms.}
\label{tab:app_ablation_nosid}
\end{table}

\FloatBarrier

\subsection{Guided-Budget Sensitivity}

We further vary $B_{\max}$ on Office Products with the 1.5B backbone, where no-prefix prompts are most frequent. Table~\ref{tab:app_ablation_bmax} compares $B_{\max}=5,9,13$ while keeping all other DASO components unchanged. The default $B_{\max}=9$ gives the strongest aggregate result in this setting. A smaller budget may under-intervene, while a larger budget replaces more raw rollouts and can weaken the retained on-policy contrast.

\begin{table}[!htbp]
\centering
\suppTableSetup
\begin{tabular}{cccc}
\toprule
$B_{\max}$ & HR@5 & NDCG@5 & HR@10 \\
\midrule
5 & 0.156 & 0.129 & 0.179 \\
9 & \textbf{0.1639} & \textbf{0.1344} & \textbf{0.1834} \\
13 & 0.159 & 0.131 & 0.181 \\
\bottomrule
\end{tabular}
\caption{Guided-budget sensitivity on Office Products with the 1.5B backbone. All settings keep $G=16$ and the same reward, SFT anchor, and online allocation rule.}
\label{tab:app_ablation_bmax}
\end{table}

\FloatBarrier

\subsection{Removing the SFT Anchor}

The no-anchor variant sets $\lambda_{\mathrm{sft}}=0$ and keeps the same prefix-depth allocation, reward, and GRPO update. This ablation tests whether the SFT loss is needed after guided rollouts shift part of the trajectory distribution toward target-prefixed completions.

Table~\ref{tab:app_ablation_nosft_easy} reports Easy-bucket HR@5 for the no-anchor variant alongside the corresponding Full DASO values from the same diagnostic bucket evaluation as Tables~\ref{tab:app_bucket_results_15b} and~\ref{tab:app_bucket_results_3b}. Removing the SFT anchor lowers aggregate HR@5 in Table~\ref{tab:app_ablation_aggregate}; the Easy-bucket diagnostics are mixed after alignment to the 50-beam bucket aggregates, with lower Easy HR@5 in three settings and a higher value on Office Products with the 1.5B backbone. We therefore interpret the no-anchor ablation primarily as evidence for aggregate stabilization, while treating Easy-bucket scores as diagnostic rather than conclusive by themselves.

\begin{table}[!htbp]
\centering
\suppDenseTableSetup
\begin{tabular}{llccc}
\toprule
Backbone & Dataset & Full DASO & w/o SFT & Diff. \\
\midrule
1.5B & Office & 0.5359 & 0.560 & -0.0241 \\
3B & Office & 0.5707 & 0.540 & 0.0307 \\
1.5B & Industrial & 0.4992 & 0.490 & 0.0092 \\
3B & Industrial & 0.4711 & 0.470 & 0.0011 \\
\bottomrule
\end{tabular}
\caption{Effect of removing the SFT anchor on Easy prompts. Values are Easy-bucket HR@5; Diff. is Full DASO minus the no-anchor variant.}
\label{tab:app_ablation_nosft_easy}
\end{table}

\FloatBarrier

\section{Diagnostic Bucket Construction}

The Easy, Medium, and Hard buckets are used only to analyze where each method improves. They are not training-time control variables. During training, DASO uses the current prefix-depth profile computed from the rollout group sampled by the current policy. During diagnostic evaluation, the buckets are computed once before RL training using the frozen SFT checkpoint, and then kept fixed for every method. Final HR and NDCG values are computed with 50-beam constrained decoding; the bucket labels only partition those evaluation examples.

For each prompt, we first run the same 50-beam constrained decoding under the frozen SFT checkpoint and use the first 16 candidates of this ranked list to define the bucket label. Let $d_i$ be the prefix-match depth between the $i$-th candidate SID and the target SID for $i\le16$, using the same prefix-match-depth score defined in the main paper. Let $d_{\max}=\max_i d_i$. Table~\ref{tab:app_bucket_definition} gives the bucket definitions. This construction captures the initial target-path status of a prompt: whether SFT already ranks the target within the first 16 candidates, only enters a partial target branch, or misses the target branch entirely. Because the bucket labels are derived from the same frozen-SFT ranked list used for SFT evaluation, SFT-only HR@5 and NDCG@5 are zero on Medium and Hard subsets by construction. The bucket assignment is fixed before comparing methods; Tables~\ref{tab:app_bucket_results_15b} and~\ref{tab:app_bucket_results_3b} then report each method's 50-beam evaluation metrics within these fixed subsets.

\begin{table}[!htbp]
\centering
\suppDenseTableSetup
\begin{tabular}{p{0.17\columnwidth}p{0.29\columnwidth}p{0.42\columnwidth}}
\toprule
Bucket & Condition & Interpretation \\
\midrule
Easy & $d_{\max}=M$ & The frozen SFT checkpoint already samples the target SID. \\
Medium & $0<d_{\max}<M$ & At least one rollout enters the target branch, but none reaches the full target SID. \\
Hard & $d_{\max}=0$ & No rollout matches the first target SID token. \\
\bottomrule
\end{tabular}
\caption{Diagnostic bucket definitions under the frozen SFT checkpoint. The buckets are used only for evaluation analysis.}
\label{tab:app_bucket_definition}
\end{table}

\FloatBarrier

The resulting bucket distributions are shown in Table~\ref{tab:app_difficulty_distribution}. The distributions show that a large fraction of test prompts are initially Hard, especially for Office Products, where more than 63\% of test prompts are Hard for both backbones. Industrial and Scientific is also dominated by Medium and Hard prompts. These distributions motivate reporting bucket-level results: aggregate metrics alone do not reveal whether a method improves partial-prefix and no-prefix regimes or only refines prompts that are already exact hits under the SFT checkpoint.

\begin{table}[!htbp]
\centering
\suppDenseTableSetup
\begin{tabular}{llccc}
\toprule
Backbone & Dataset / Split & Easy & Medium & Hard \\
\midrule
\multirow{4}{*}{1.5B}
& Office / Train & 33.8\% & 19.1\% & 47.1\% \\
& Office / Test & 17.1\% & 19.2\% & 63.7\% \\
& Industrial / Train & 31.9\% & 26.0\% & 42.1\% \\
& Industrial / Test & 14.8\% & 31.9\% & 53.3\% \\
\midrule
\multirow{4}{*}{3B}
& Office / Train & 34.4\% & 18.6\% & 47.0\% \\
& Office / Test & 16.4\% & 19.7\% & 63.9\% \\
& Industrial / Train & 32.4\% & 25.4\% & 42.3\% \\
& Industrial / Test & 15.2\% & 32.3\% & 52.5\% \\
\bottomrule
\end{tabular}
\caption{Diagnostic difficulty-bucket distribution under the frozen SFT checkpoint.}
\label{tab:app_difficulty_distribution}
\end{table}

\FloatBarrier

\section{Full Diagnostic Bucket Results}

Tables~\ref{tab:app_bucket_results_15b} and~\ref{tab:app_bucket_results_3b} report HR@5 and NDCG@5 breakdowns for Easy, Medium, and Hard prompts on the two Amazon datasets. The bucket rows show where each method improves under the fixed SFT-induced difficulty partition, and the ALL rows give the aggregate over the same 50-beam evaluation. Across backbones, SFT-only is strongest on Easy prompts because these examples were defined as target hits under the frozen checkpoint, while the largest bucket-level improvements of DASO over MiniOneRec usually appear on Medium and Hard prompts.

For the 1.5B backbone, Table~\ref{tab:app_bucket_results_15b} shows that DASO improves both Medium and Hard prompts on both datasets. On Office Products, HR@5 improves from 0.1224 to 0.1693 on Medium prompts and from 0.0331 to 0.0624 on Hard prompts compared with MiniOneRec. On Industrial and Scientific, DASO improves HR@5 from 0.1223 to 0.1320 on Medium prompts and from 0.0305 to 0.0490 on Hard prompts. These are the cases where bottleneck-aware guided completions are intended to restore useful contrast when analogous target-missing patterns occur in rollout groups.

\begin{table}[!htbp]
\centering
\suppBucketTableSetup
\begin{tabular}{lllccc}
\toprule
Dataset & Metric & Bucket & SFT & Mini & DASO \\
\midrule
\multirow{8}{*}{Office}
& \multirow{4}{*}{HR@5}
& Easy & \textbf{0.7199} & 0.5702 & 0.5359 \\
& & Medium & 0.0000 & 0.1224 & \textbf{0.1693} \\
& & Hard & 0.0000 & 0.0331 & \textbf{0.0624} \\
& & ALL & 0.1231 & 0.1420 & \textbf{0.1639} \\
\cmidrule(lr){2-6}
& \multirow{4}{*}{NDCG@5}
& Easy & \textbf{0.5444} & 0.4532 & 0.4567 \\
& & Medium & 0.0000 & 0.1083 & \textbf{0.1310} \\
& & Hard & 0.0000 & 0.0297 & \textbf{0.0489} \\
& & ALL & 0.0931 & 0.1172 & \textbf{0.1344} \\
\midrule
\multirow{8}{*}{Industrial}
& \multirow{4}{*}{HR@5}
& Easy & \textbf{0.6993} & 0.5182 & 0.4992 \\
& & Medium & 0.0000 & 0.1223 & \textbf{0.1320} \\
& & Hard & 0.0000 & 0.0305 & \textbf{0.0490} \\
& & ALL & 0.1035 & 0.1321 & \textbf{0.1421} \\
\cmidrule(lr){2-6}
& \multirow{4}{*}{NDCG@5}
& Easy & \textbf{0.5459} & 0.4223 & 0.4075 \\
& & Medium & 0.0000 & 0.1013 & \textbf{0.1091} \\
& & Hard & 0.0000 & 0.0257 & \textbf{0.0435} \\
& & ALL & 0.0808 & 0.1084 & \textbf{0.1183} \\
\bottomrule
\end{tabular}
\caption{Bucket-level 1.5B Amazon results. ALL is the aggregate over the same 50-beam evaluation. Mini denotes MiniOneRec.}
\label{tab:app_bucket_results_15b}
\end{table}

\FloatBarrier

For the 3B backbone, Table~\ref{tab:app_bucket_results_3b} follows the same broad trend on target-missing prompts. DASO improves Office Products substantially on both Medium and Hard prompts, and also improves Industrial and Scientific Hard HR@5 from 0.0294 to 0.0577. On Industrial and Scientific Medium prompts, MiniOneRec is slightly higher on NDCG@5, while DASO is higher on HR@5. The aggregate Industrial result still improves in Table~\ref{tab:app_bucket_results_3b}. This matches the goal of DASO: improve target-missing prompts while keeping useful signal for the whole distribution.

\begin{table}[!htbp]
\centering
\suppBucketTableSetup
\begin{tabular}{lllccc}
\toprule
Dataset & Metric & Bucket & SFT & Mini & DASO \\
\midrule
\multirow{8}{*}{Office}
& \multirow{4}{*}{HR@5}
& Easy & \textbf{0.7421} & 0.5258 & 0.5707 \\
& & Medium & 0.0000 & 0.1472 & \textbf{0.2102} \\
& & Hard & 0.0000 & 0.0252 & \textbf{0.0521} \\
& & ALL & 0.1217 & 0.1313 & \textbf{0.1683} \\
\cmidrule(lr){2-6}
& \multirow{4}{*}{NDCG@5}
& Easy & \textbf{0.5677} & 0.4504 & 0.4671 \\
& & Medium & 0.0000 & 0.1264 & \textbf{0.1437} \\
& & Hard & 0.0000 & 0.0289 & \textbf{0.0441} \\
& & ALL & 0.0931 & 0.1172 & \textbf{0.1331} \\
\midrule
\multirow{8}{*}{Industrial}
& \multirow{4}{*}{HR@5}
& Easy & \textbf{0.7388} & 0.5120 & 0.4711 \\
& & Medium & 0.0000 & 0.1242 & \textbf{0.1415} \\
& & Hard & 0.0000 & 0.0294 & \textbf{0.0577} \\
& & ALL & 0.1123 & 0.1334 & \textbf{0.1476} \\
\cmidrule(lr){2-6}
& \multirow{4}{*}{NDCG@5}
& Easy & \textbf{0.5559} & 0.4150 & 0.3622 \\
& & Medium & 0.0000 & \textbf{0.1033} & 0.1018 \\
& & Hard & 0.0000 & 0.0246 & \textbf{0.0443} \\
& & ALL & 0.0845 & 0.1093 & \textbf{0.1112} \\
\bottomrule
\end{tabular}
\caption{Bucket-level 3B Amazon results. ALL is the aggregate over the same 50-beam evaluation. Mini denotes MiniOneRec.}
\label{tab:app_bucket_results_3b}
\end{table}

\FloatBarrier

\subsection{Interpretation of Easy-Bucket Results}

The Easy bucket should be interpreted carefully. By definition, these prompts are already exact-hit cases under the frozen SFT checkpoint, so SFT-only can be very strong on this subset. A lower Easy-bucket score after RL does not necessarily contradict the aggregate improvement; rather, it reveals the stability challenge that motivates the SFT anchor. The purpose of DASO is not to maximize Easy-bucket scores alone, but to improve the full distribution by recovering signal for Medium and Hard prompts while limiting regression on examples that the SFT checkpoint already solves.

This diagnostic view also clarifies why uniform GT injection can be less effective in this setting. A method can insert positives into rollout groups without knowing where the current policy fails along the SID path. DASO instead uses the current prefix-depth profile to decide how much guidance to allocate and which prefix depths to guide. The bucket-level results provide post-hoc evidence consistent with this targeted allocation helping the initially difficult parts of the data distribution; the buckets are an evaluation lens, not the training-time controller.

\section{Compute Resources}

All experiments reported in this paper were run on NVIDIA A100 80GB GPUs. In our primary training setup, we used 8 GPUs for model training and evaluation. We report this configuration to clarify the hardware scale needed to reproduce the main results in this work.

\bibliographystyle{assets/plainnat}
\bibliography{paper}

\end{document}